\PassOptionsToPackage{table}{xcolor}
\documentclass{article} 
\usepackage[T1]{fontenc} 
\usepackage{iclr2026_conference,times}

\usepackage{amsmath,amsfonts,bm}

\def\eqref#1{equation~\ref{#1}}

\def\1{\bm{1}}

\DeclareMathAlphabet{\mathsfit}{\encodingdefault}{\sfdefault}{m}{sl}
\SetMathAlphabet{\mathsfit}{bold}{\encodingdefault}{\sfdefault}{bx}{n}

\usepackage{hyperref}
\usepackage{url}
\usepackage{adjustbox}
\usepackage{booktabs}
\usepackage{xspace}
\usepackage{amsmath}
\usepackage{amssymb}
\usepackage{amsthm}
\usepackage{algorithm}
\usepackage{algpseudocode}
\usepackage{xcolor}
\usepackage{tcolorbox}
\usepackage{booktabs}
\usepackage{adjustbox}
\usepackage{bm}
\usepackage{opd_frontmatter}

\newtheorem{proposition}{Proposition}[section]
\newtheorem{theorem}{Theorem}[section]

\definecolor{methodshade}{RGB}{235,243,250}

\title{Which Successes Make Good References? Action-Sequence Alignment for Multi-Turn On-Policy Distillation}

\author{
Mufei Li\textsuperscript{1, *}, 
Nikki Lijing Kuang\textsuperscript{2},
Zhuoer Wang\textsuperscript{2},
Samson Koelle\textsuperscript{2},
Jie Yuan\textsuperscript{2},
Manqing Mao\textsuperscript{2},
Hong Wang\textsuperscript{2},
Yanjun Lin\textsuperscript{2},
James Feng\textsuperscript{2},
Daniel Edmiston\textsuperscript{2},
Pan Li\textsuperscript{1}, 
Wei Niu\textsuperscript{2}
}
\affiliations{\textsuperscript{1}Georgia Institute of Technology,
  \textsuperscript{2}Amazon Web Services}
\contact{%
  \emailgroup{mufei.li,panli}{gatech.edu}\\
  \emailgroup{nikkiklj,wze,niuwei}{amazon.com}%
}
\institutionlogos{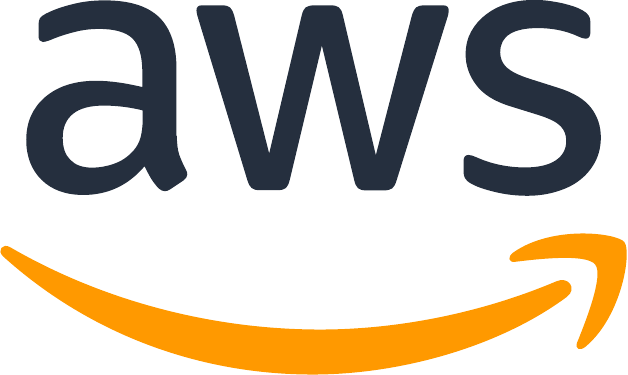}{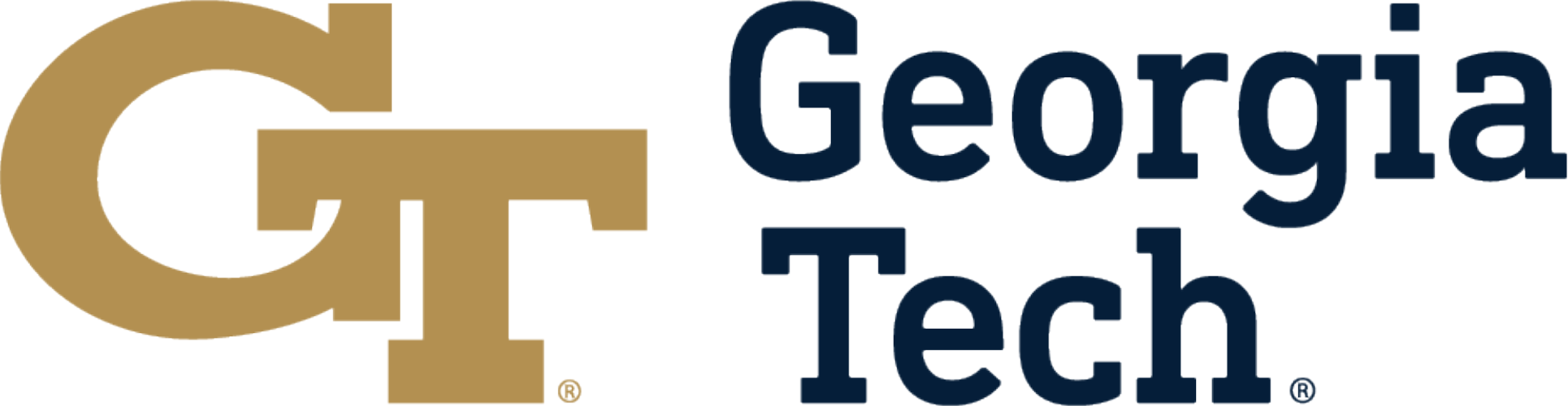}

\iclrfinalcopy 
\begin{document}

\begin{frontmatter}
\maketitle

\begin{abstract}
On-policy distillation (OPD) reduces train-inference distribution mismatch by supervising a student on its own trajectories. For multi-turn tool-use agents, reference-augmented OPD further conditions the teacher on a successful trajectory from the same task, providing hindsight on how to complete the task. However, not all successful trajectories make good references. Different successful trajectories may follow different interaction paths and reach incompatible states, providing misleading supervision for a failed student trajectory. We formulate successful-reference selection as a free-target optimal transport problem, which reduces to assigning each failed rollout its most compatible successful reference. Since latent environment states are hidden, we use action-sequence alignment as an observable proxy for compatibility and instantiate it with longest common subsequence (LCS), which identifies a shared ordered backbone of actions while allowing recoverable detours. This makes trajectory compatibility the central criterion for reference selection in multi-turn OPD. Experiments on ALFWorld and the airline domain of $\tau^3$-bench show that LCS is the only evaluated reference selector to improve over reference-free vanilla OPD on both ALFWorld overall scores and all three airline metrics. The resulting airline-trained models also transfer to the retail domain, where LCS improves all three metrics.
\end{abstract}
\end{frontmatter}

\begingroup
  \renewcommand{\thefootnote}{\fnsymbol{footnote}}%
  \footnotetext[1]{Work done during an internship at Amazon Web Services.}%
\endgroup

\section{Introduction}

By supervising a student on its own trajectories, on-policy distillation (OPD) mitigates train-inference distribution mismatch~\citep{agarwal2024onpolicy} and has been adopted in the post-training of frontier models~\citep{yang2025qwen3, cursor2026composer25, zai2026glm52, kimiteam2026kimik3}. However, extending OPD to multi-turn tool-use agents introduces additional challenges. These agents often operate over long horizons under partial observability: user goals, database states, and environment layouts may be revealed only incrementally through dialogue, API calls, and interaction~\citep{yao2024tau,shridhar2021alfworld}. Their decisions therefore exhibit strong sequential dependencies, as earlier actions reveal information, alter the environment, and constrain the states that remain reachable and the actions that remain valid. Consequently, a local error can redirect the subsequent trajectory and compound across turns.

This motivates reference-augmented OPD, which uses successful trajectories to guide teacher supervision. Specifically, the teacher scores a student trajectory conditioned on a reward-validated successful trajectory from the same task instance~\citep{yu2026multi, li2026distill, hubotter2026reinforcement}. 
We call this complete successful interaction trajectory a \textbf{reference trajectory}, or simply a \textbf{reference}. It captures a feasible sequence of environment observations and executed actions, which is provided to the teacher only as training-time hindsight.
By exposing the outcomes of a feasible interaction path, the reference may help the teacher diagnose errors and provide more informative supervision. Yet a task may admit multiple successful trajectories following different paths. It remains unclear whether they are equally useful for correcting a given student trajectory. Indeed, recent work finds that conditioning on a successful trajectory can even degrade multi-turn distillation~\citep{liu2026hero}. This raises our central question: \emph{Which successes make good references for multi-turn OPD?}

\begin{figure*}[t]
    \centering
    \includegraphics[
    width=\linewidth,
    ]{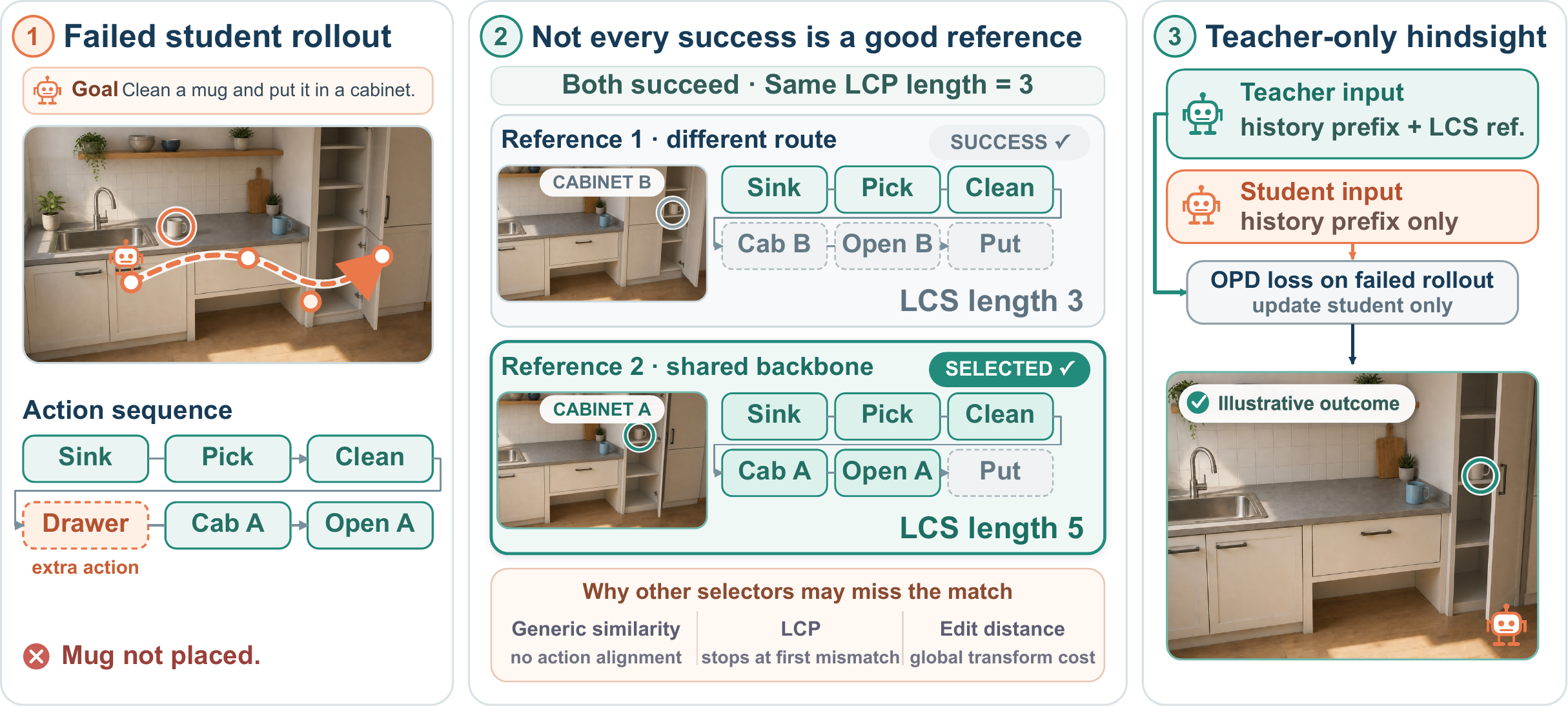}
\caption{\textbf{LCS selects compatible references for failure-only OPD.} Two same-task successes have the same LCP length (3), but LCS favors the reference with greater ordered action agreement (5 vs.\ 3). The reference conditions only the teacher; the student sees only its own history.}
    \label{fig:teaser}
    \vspace{-0.3cm}
\end{figure*}

We propose to select successful references based on their action-sequence alignment with each failed student trajectory. We formalize this selection as a free-target optimal transport problem, which reduces to assigning each failed trajectory its most compatible same-task success. This choice is motivated by two properties of multi-turn interaction: path dependence and the connection between action sequences and latent state paths.
 
Since earlier actions shape subsequent states and valid actions, reference selection must consider interaction paths rather than isolated turns. For example, an early database write can change the records returned by later queries and the validity of subsequent API calls. Similarly, moving an object can alter the future room configuration in embodied agentic tasks. Two turns that appear locally similar may therefore correspond to incompatible states.

Given this path dependence, reference selection would ideally compare the latent state paths of failed and successful trajectories. Although these paths are hidden under partial observability, they evolve according to each trajectory's actions. Suppose two trajectories share the same initial state and the environment transition is deterministic given the current state and a fully specified action. Identical action prefixes then certify identical state prefixes, motivating action-sequence alignment as an observable proxy for trajectory compatibility.

Although the state-path connection above
naturally motivates longest common prefix (LCP), exact prefix matching
discards all agreement after the first unmatched action and can be overly
restrictive when that action is redundant or part of a recoverable detour. We therefore use longest common subsequence (LCS) as a relaxed measure of compatibility. Unlike LCP, LCS retains ordered action agreement beyond the first mismatch, allowing it to identify shared action subsequences despite unmatched actions in between.
The normalized LCS score then instantiates the pairwise compatibility
function in the optimal-transport framework. Fig.~\ref{fig:teaser} summarizes the resulting reference-selection
and distillation pipeline.

We evaluate LCS on ALFWorld and the airline domain of $\tau^3$-bench using two student models per benchmark, against seven alternative reference selectors. Across both benchmarks, LCS is the only evaluated selector to improve over vanilla OPD on every aggregate metric for both students, while consistently outperforming LCP, its strict prefix-based counterpart. On ALFWorld, LCS achieves the highest overall success rates on both Valid Seen and Valid Unseen for both students. Without further optimization, airline-trained models also transfer to the retail domain, where LCS improves all three metrics for both students. These results show that successful references are not interchangeable: trajectory compatibility matters for multi-turn OPD.
\section{Background and Problem Formulation}

\label{sec:back}

\textbf{POMDP.} We formulate multi-turn tool-use interaction as a partially observable Markov decision process (POMDP), in which the agent observes only partial information about the latent environment state~\citep{KAELBLING199899}. At trajectory turn $t\in\{0,\cdots, T-1\}$, the agent observes $o_t$, which provides partial information about the latent state $s_t$. It then generates a response $u_t=(c_t, a_t)$, where $c_t$ is a chain-of-thought reasoning trace (e.g., $\langle \text{think} \rangle\cdots \langle /\text{think} \rangle$), and $a_t$ is an action. Executing $a_t$ yields the next observation $o_{t+1}$. Because $s_t$ is not directly observable, the agent conditions its response $u_t$ on the full interaction history:
$
    h_t=(o_0,a_0,\cdots, o_{t-1}, a_{t-1}, o_t)$.

\textbf{On-Policy Distillation (OPD).} 
Given a teacher policy $\pi_\phi$ and a student policy
$\pi_\theta$, OPD minimizes reverse KL divergence on
student-visited trajectories:
$
    \mathcal{L}(\theta)=\mathbb{E}_{h_t\sim \pi_{\theta}}[\mathcal{D}_{KL}(\pi_{\theta}(\cdot\mid h_t) \| \pi_{\phi}(\cdot\mid h_t))]
$.
In practice, this objective is often approximated with 1-sample Monte Carlo estimation:
$
    \mathbb{E}_{(h_{0:T-1},u_{0:T-1})\sim\pi_{\theta}}
    \left[
        \frac{1}{T}\sum_{t=0}^{T-1}
        \log\frac{\pi_{\theta}(u_t\mid h_t)}
                  {\pi_{\phi}(u_t\mid h_t)}
    \right]$.

\textbf{Reference-Augmented OPD.} Prior work augments distillation by conditioning the teacher on
a validated successful trajectory generated by the teacher for the same task~\citep{yu2026multi, hubotter2026reinforcement}. A trajectory $\tau=(o_0,a_0,\ldots,o_{T-1},a_{T-1},o_T)$
records observations and executed actions. Given a successful reference \(\tau\), the teacher scores the student-generated response \(u_t\) using \(\pi_\phi(u_t\mid h_t,\tau)\), where \(h_t\) is the history of the student rollout. The student remains conditioned only on its own history, using \(\pi_\theta(u_t\mid h_t)\), and therefore requires no reference at inference.

\textbf{Reference Selection.} A substantial fraction of agentic tasks admit multiple successful interaction trajectories. In our empirical studies, among training tasks with at least one reward-validated success, $33\%$ on ALFWorld and $82.8\%$ on $\tau^3$-bench airline have at least two distinct successful trajectories across four teacher rollouts (Appendix Table~\ref{tab:reference_pool_distribution}). These successes need not be equally useful as references for a given failed student trajectory. Here, we study selection among successful interaction trajectories for reference-augmented multi-turn OPD.

For a task $x$, let $\mathcal{R}_x$ denote its set of reward-validated
successful trajectories. Given a failed student trajectory $\tau^-$ for
$x$, a reference selector $f$ returns
    $\tau^* = f(x,\tau^-) \in \mathcal{R}_x$,
where $\tau^*$ is provided to the teacher as the reference when scoring
$\tau^-$.

\section{Methodology}

\subsection{An Optimal Transport Perspective on Reference Selection}
\label{sec: OT_select}

\begin{figure}[t]
    \centering
    \includegraphics[width=0.6\linewidth]{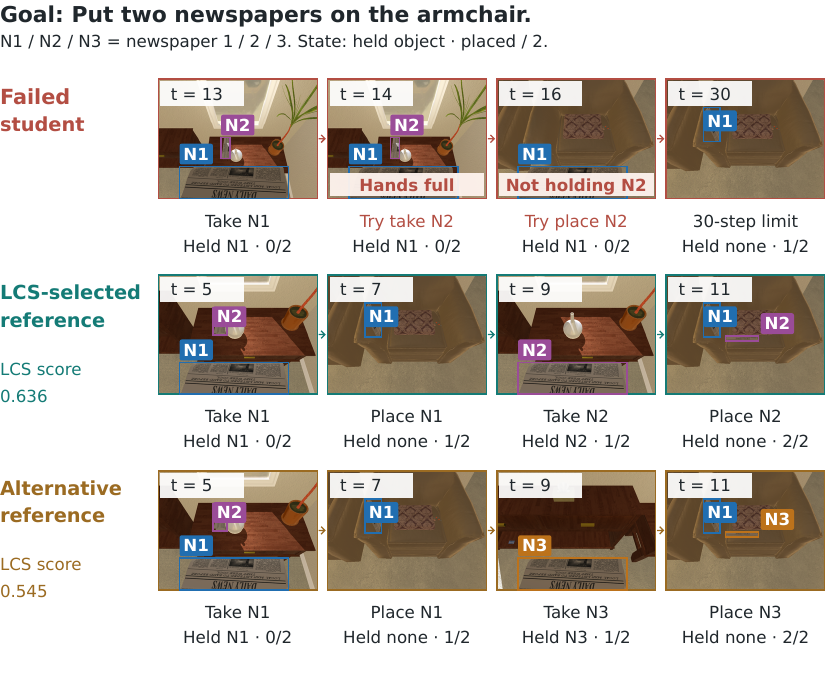}
    \caption{\textbf{Reference selection in an ALFWorld task.}
Both successful references place N1, then N2 or N3. LCS selects the one with greater ordered action agreement with the failed rollout. Only key turns are shown; LCS scores are computed over the full action sequences.}
    \label{fig:newspaper}
    \vspace{-0.3cm}
\end{figure}

\textbf{Reference Utility through Trajectory Compatibility.} Although all trajectories in $\mathcal{R}_x$ complete task $x$, they vary in usefulness for correcting a given failed trajectory \(\tau^-\). A reference preserving the student's valid choices may help the teacher isolate errors, whereas one following a different path may require broader corrections and risk disrupting correct behavior. For example, in Fig.~\ref{fig:newspaper}, the student attempts to take N2 while holding N1. The N1/N2 reference preserves the choice of N2, whereas the N1/N3 reference instead takes N3. We therefore model reference utility with a directional pairwise correction cost, where lower costs indicate less correction needed to reconcile the failed trajectory with the reference.

\textbf{Reference-Guided Correction as Optimal Transport (OT).} Aggregating this pairwise correction cost across failed rollouts naturally yields an optimal transport formulation~\citep{47813}. For task $x$, the empirical
distribution over failed trajectories forms the source distribution, and
the successful trajectories in $\mathcal{R}_x$ serve as candidate targets.
We seek a transport plan that assigns source mass to these targets while minimizing the aggregate correction cost.

\textbf{Empirical Source and Pairwise Cost.}
For a task $x$ with $\mathcal{R}_x\neq\varnothing$, let $\{\tau_i^-\}_{i=1}^{N_x}$ be its $N_x\geq 1$ failed on-policy rollouts, represented by the uniform
empirical distribution
$
    \mu_x^-
    =
    \frac{1}{N_x}
    \sum_{i=1}^{N_x}
    \delta_{\tau_i^-},
$
where $\delta_{\tau}$ denotes a unit point mass at $\tau$.
Let $c(\tau^-,\tau^+)\in\mathbb{R}$ denote the directional cost of assigning a failed
trajectory $\tau^-$ to a successful reference
$\tau^+\in\mathcal{R}_x$. Given a compatibility score $s(\tau^-,\tau^+) \in \mathbb{R}$, we set $
    c(\tau^-,\tau^+)
    =
    -s(\tau^-,\tau^+)$, so minimizing cost maximizes compatibility.

\textbf{Free-Target Optimal Transport.} Let $\mathcal{P}(\mathcal{R}_x)$ denote the probability measures supported on $\mathcal{R}_x$. We formulate reference selection as the following free-target optimal transport problem:
\begin{equation}
\begin{aligned}
    \mathsf{OT}_{c}(\mu_x^-,\mathcal{R}_x)
    &=
    \min_{\nu\in\mathcal{P}(\mathcal{R}_x)}
    \min_{\gamma\in\Pi(\mu_x^-,\nu)}
    \int
    c(\tau^-,\tau^+)
    \,\mathrm{d}\gamma(\tau^-,\tau^+),
\end{aligned}
    \label{eq:free_target_ot}
\end{equation}
where $\Pi(\mu_x^-,\nu)$ denotes transport plans with marginals $\mu_x^-$ and $\nu$. Unlike standard optimal transport, we optimize the target marginal $\nu$ jointly, allowing reference usage to adapt to pairwise compatibility without fixed selection frequencies.

\begin{proposition}[Optimal Reference Assignment]
\label{prop:optimal_reference_assignment}
Suppose that $\mathcal{R}_x$ is finite and nonempty,
with finite pairwise costs.
Then the free-target OT problem in
Equation~\ref{eq:free_target_ot}
has an optimal value equal to the average minimum
pairwise cost across failed rollouts.
Moreover, it admits an optimal deterministic plan
that independently assigns each failed rollout
to a minimum-cost reference.
\end{proposition}

The proof is provided in
Appendix~\ref{appendix:proof-prop-optimal-reference}.
Since $c=-s$, the proposition instantiates the
reference selector $f$ from Section~\ref{sec:back} as
$
    f(x,\tau^-)
    \in
    \arg\max_{\tau^+\in\mathcal{R}_x}
    s(\tau^-,\tau^+).
$
Thus, free-target OT reduces to independent
nearest-reference assignment.

\textbf{Training Implication: Failure-Only OPD.}
The nearest-reference rule assigns successful
references to failed student rollouts, reflecting a
correction-oriented view of OPD. This motivates restricting OPD updates to failed rollouts. Student successes have already achieved the desired task outcome, even when their trajectories
differ from those in $\mathcal{R}_x$, and further teacher supervision may introduce unnecessary corrections or disrupt valid behavior, particularly when the teacher makes mistakes.
Prior work also supports failure-only distillation:
SRPO finds it retains most benefits, while success-only
distillation degrades performance and destabilizes
training~\citep{li2026unifying};
PTD-PO further reports that failure-only distillation
outperforms distillation on all
trajectories~\citep{xiang2026teaching}.
We therefore apply OPD only to failed student rollouts.

\subsection{Action-Sequence Alignment}
\label{sec:action_sequence_align}

\textbf{Latent-State Compatibility.} The OT formulation leaves a key choice: how to measure trajectory compatibility. For a trajectory
$\tau=(o_0,a_0,\ldots,o_{T-1},a_{T-1},o_T)$, let
$\mathbf{s}(\tau)=(s_0,\ldots,s_T)$ denote its latent state path. Ideally,
compatibility would compare $\mathbf{s}(\tau^-)$ and
$\mathbf{s}(\tau^+)$, but these paths are hidden under partial observability.

\textbf{Why Not Align Observation Sequences?}
A natural alternative is to compare observation sequences
$\mathbf{o}(\tau)=(o_0,\ldots,o_T)$.
However, observations reveal only partial state information
and are often long, free-form, and sensitive to formatting.
For example, API responses may contain the same task-relevant
record but differ in field order or metadata, while
identical-looking room descriptions may conceal different
object locations or inventory contents due to earlier
actions~\citep{shridhar2021alfworld,yao2024tau}.
Thus, compatible states may produce dissimilar observations,
while similar observations may conceal incompatible states,
making observation overlap an unreliable proxy for
state-path compatibility.

\textbf{Proposal: Action-Sequence Alignment.}
We instead compare executed action sequences
$\mathbf{a}(\tau)=(a_0,\ldots,a_{T-1})$.
Tool calls consist of a tool name and a few
arguments, making them easier to compare than
observations~\citep{NEURIPS2023_d842425e}.
More importantly, actions are agent-controlled inputs
to state transitions: they reveal information, modify
the environment, and constrain reachable states.
Thus, action-sequence alignment provides an observable
proxy for latent state-path compatibility.

\textbf{State-Path Interpretation.}
For trajectories sharing an initial state and the same
deterministic state-transition dynamics, identical
action prefixes imply identical state prefixes. Under stochastic transitions, an identical action prefix instead induces the same distribution over state prefixes when the initial-state distribution is shared.

\label{sec:relax_exact_prefix}

\textbf{Limitations of Exact Prefix Matching.} The state-path interpretation naturally motivates longest common prefix (LCP) as a conservative compatibility criterion. However, LCP discards all agreement after the first unmatched action,
implicitly treating that mismatch as the end of the trajectories' usable
shared path. This can be overly restrictive because an unmatched action may be redundant, may fail without changing the environment, or may be part of a detour whose effects are later reversed.

\textbf{Longest Common Subsequence (LCS) Relaxation.}
We relax exact prefix matching using LCS.
Given $\mathbf{a}(\tau^-)=(a_0^-,\ldots,a_{T^--1}^-)$
and $\mathbf{a}(\tau^+)=(a_0^+,\ldots,a_{T^+-1}^+)$,
their LCS length is
\begin{equation*}
\ell_{\mathrm{LCS}}(\tau^-,\tau^+)
=
\max
\left\{
m
\;\middle|\;
\substack{
0 \leq i_1 < \cdots < i_m < T^-,\\
0 \leq j_1 < \cdots < j_m < T^+,\\
a^-_{i_r}=a^+_{j_r},\quad r=1,\ldots,m
}
\right\}.
\end{equation*}
LCS preserves action order while allowing unmatched actions,
capturing agreement beyond the first mismatch and distinguishing
references with identical LCP scores
(Fig.~\ref{fig:teaser}).

\textbf{Normalized LCS Compatibility.}
To account for trajectory length differences, we define
\begin{equation*}
    s_{\mathrm{LCS}}(\tau^-,\tau^+)
    =
    \frac{\ell_{\mathrm{LCS}}(\tau^-,\tau^+)}
    {Z(T^-,T^+)},
\end{equation*}
where $Z(T^-,T^+)>0$ is a normalizer.
For example, $Z=T^-$ measures failed-trajectory coverage,
while $Z=\min\{T^-,T^+\}$ measures agreement relative to
the shorter trajectory.
This score instantiates the compatibility function
in Sec.~\ref{sec: OT_select}.

With all components in place, we summarize the
resulting training procedure.
We first collect a fixed number of teacher rollouts
per task offline and retain reward-validated successes
as candidate references.
During online OPD, we update the student only on
failed rollouts, conditioning the teacher on the
highest-scoring same-task reference when available;
otherwise, we use standard reference-free OPD.
The complete algorithm is provided in
Algorithm~\ref{alg:lcs-ref-opd} in the Appendix.

\subsection{Theoretical Guarantees}

We give sufficient conditions under which LCS matches preserve latent-state consistency and compatible references improve teacher supervision.

\begin{theorem}[State Consistency under Recoverable Detours]
\label{thm:lcs_recoverable_detours}
For a failed trajectory $\tau^-$ of task $x$ and a
reference $\tau^+\in\mathcal R_x$, let
$m=\ell_{\mathrm{LCS}}(\tau^-,\tau^+)$ and fix an LCS
alignment
\begin{equation*}
A_{\mathrm{LCS}}(\tau^-,\tau^+)
=
\bigl((i_r,j_r)\bigr)_{r=1}^{m},
\label{eq:lcs_alignment}
\end{equation*}
where $0\leq i_1<\cdots<i_m<T^-$,
$0\leq j_1<\cdots<j_m<T^+$, and
$a_{i_r}^-=a_{j_r}^+$ for all $r=1,\ldots,m$.
Let $s_t^\pm$ denote the latent states before
actions $a_t^\pm$.

Suppose the trajectories share an initial state and
follow the same deterministic dynamics
$s_{t+1}=F(s_t,a_t)$.
Assume that, before each matched pair, the unmatched action blocks following the preceding match (or the trajectory start) reach the same state whenever entered from the same state (recoverability).
Then every LCS-matched pair is
\emph{state-consistent}:
$s_{i_r}^-=s_{j_r}^+$ for all $r=1,\ldots,m$.
\end{theorem}

This condition allows different intermediate
deviations that reconverge before the next matched
action, rather than requiring every unmatched action
to have no effect. This formalizes the intuition in Sec.~\ref{sec:relax_exact_prefix} that LCS can retain meaningful
post-mismatch agreement when an early divergence is redundant, ineffective, or later reversed. The proof
is provided in
Appendix~\ref{appendix:proof_lcs_recoverable_detours}.

\textbf{When Does a Compatible Reference Improve Supervision?}
State consistency alone does not guarantee improved
teacher supervision. Let $q_t^\ast(\cdot\mid h_t)$
denote an ideal corrective target distribution at history $h_t$, defined independently of the reference choice.
For a candidate reference $\tau^+$, define its
supervision gain at turn $t$ as
\begin{equation*}
\Delta_t(\tau^+)
=
\mathbb{E}_{u_t\sim q_t^\ast(\cdot\mid h_t)}
\left[
\log
\frac{
\pi_\phi(u_t\mid h_t,\tau^+)
}{
\pi_\phi(u_t\mid h_t)
}
\right].
\label{eq:teacher_gain}
\end{equation*}
Positive gain means that reference conditioning assigns
higher expected log probability to desirable responses
than reference-free supervision.

\begin{theorem}[Benefit over Reference-Free Supervision]
\label{thm:reference_supervision_gain}
Under the conditions of
Theorem~\ref{thm:lcs_recoverable_detours}, suppose
there exist $g>0$ and $b\geq0$ such that reference
conditioning yields total supervision gain at least $gm$
over the $m$ LCS-matched turns and at least
$-b(T^--m)$ over the remaining turns. Then
\begin{equation*}
\frac{1}{T^-}
\sum_{t=0}^{T^--1}
\Delta_t(\tau^+)
\geq
(g+b)\frac{m}{T^-}-b.
\end{equation*}
Consequently, the average supervision gain is
strictly positive whenever
$
\frac{m}{T^-}>\frac{b}{g+b}.
$
\end{theorem}

For fixed $g$ and $b$, greater failed-trajectory LCS
coverage strengthens this lower bound.
Appendix~\ref{appendix:proof_reference_benefit}
provides the proof.

\section{Experiments}

\begin{table*}[t]
    \centering
    \setlength{\tabcolsep}{4pt}
    \renewcommand{\arraystretch}{1.2}
    \begin{adjustbox}{width=\textwidth}
    \begin{tabular}{lllllllllllllll}
    \toprule
    & \multicolumn{7}{c}{Valid Seen Success Rate (\%)} & \multicolumn{7}{c}{Valid Unseen Success Rate (\%)} \\
    \cmidrule(lr){2-8} \cmidrule(lr){9-15}
    Approach & Pick & Clean & Heat & Cool & Examine & Pick2 & Overall & Pick & Clean & Heat & Cool & Examine & Pick2 & Overall \\
    \midrule
    Teacher Zero-Shot    & $94.3$ & $96.3$ & $87.5$ & $88.0$ & $61.5$ & $83.3$ & $87.9$ & $91.7$ & $90.3$ & $91.3$ & $90.5$ & $72.2$ & $64.7$ & $85.1$ \\
    \midrule 
    \rowcolor{gray!15}
    \multicolumn{15}{c}{
        \textbf{Qwen2.5-3B-Instruct} \textcolor{gray}{(Student)}
    } \\
    Zero-Shot    & $14.3$ & $7.4$ & $0.0$ & $4.0$ & $38.5$ & $0.0$ & $9.3$ & $29.2$ & $0.0$ & $0.0$ & $0.0$ & $38.9$ & $0.0$ & $10.5$ \\
    SFT                  & $54.3$ & $18.5$ & $18.8$ & $24.0$ & $38.5$ & $16.7$ & $30.0$ & $45.8$ & $12.9$ & $30.4$ & $19.1$ & $55.6$ & $11.8$ & $28.4$ \\
    Vanilla OPD & $94.3$ & $85.2$ & $\underline{87.5}$ & $64.0$ & $\underline{69.2}$ & $\underline{66.7}$ & $79.3$ & $87.5$ & $77.4$ & $73.9$ & $\underline{90.5}$ & $\mathbf{77.8}$ & $58.8$ & $\underline{78.4}$\\
    \midrule
    \rowcolor{gray!8}
    \multicolumn{15}{l}{\textit{Reference-Augmented OPD}} \\
    Random               & $\underline{97.1}$ & $85.2$ & $81.2$ & $\underline{76.0}$ & $38.5$ & $45.8$ & $75.0$ & $\mathbf{95.8}$ & $71.0$ & $73.9$ & $71.4$ & $27.8$ & $\underline{76.5}$ & $70.9$ \\
    Shortest             & $94.3$ & $81.5$ & $\mathbf{93.8}$ & $64.0$ & $\underline{69.2}$ & $\mathbf{70.8}$ & $\underline{80.0}$ & $87.5$ & $\mathbf{87.1}$ & $73.9$ & $66.7$ & $44.4$ & $52.9$ & $71.6$\\
    BM25                 & $94.3$ & $85.2$ & $81.2$ & $60.0$ & $53.8$ & $\underline{66.7}$ & $76.4$ & $\underline{91.7}$ & $67.7$ & $\mathbf{100.0}$ & $\underline{90.5}$ & $50.0$ & $58.8$ & $77.6$ \\
    Qwen3-Embedding-0.6B & $\mathbf{100.0}$ & $\underline{88.9}$ & $50.0$ & $60.0$ & $61.5$ & $58.3$ & $74.3$ & $87.5$ & $64.5$ & $82.6$ & $76.2$ & $\underline{61.1}$ & $58.8$ & $72.4$ \\
    Edit Distance        & $\mathbf{100.0}$ & $85.2$ & $81.2$ & $64.0$ & $30.8$ & $\underline{66.7}$ & $76.4$ & $87.5$ & $74.2$ & $69.6$ & $76.2$ & $33.3$ & $70.6$ & $70.1$ \\
    LCP                  & $\mathbf{100.0}$ & $81.5$ & $81.2$ & $68.0$ & $46.2$ & $58.3$ & $76.4$ & $87.5$ & $\underline{80.6}$ & $82.6$ & $76.2$ & $44.4$ & $64.7$ & $74.6$ \\
    BRTS                 & $91.4$ & $81.5$ & $\underline{87.5}$ & $68.0$ & $53.8$ & $58.3$ & $75.7$ & $\underline{91.7}$ & $77.4$ & $78.3$ & $\underline{90.5}$ & $50.0$ & $58.8$ & $76.1$\\
    \rowcolor{methodshade}
    LCS                  & $94.3$ & $\mathbf{92.6}$ & $\underline{87.5}$ & $\mathbf{84.0}$ & $\mathbf{84.6}$ & $\underline{66.7}$ & $\mathbf{85.7}$ & $\underline{91.7}$ & $77.4$ & $\underline{91.3}$ & $\mathbf{100.0}$ & $\underline{61.1}$ & $\mathbf{94.1}$ & $\mathbf{85.8}$ \\
    \midrule
    \rowcolor{gray!15}
    \multicolumn{15}{c}{
        \textbf{Qwen2.5-7B-Instruct} \textcolor{gray}{(Student)}
    } \\
    Zero-Shot    & $45.7$ & $14.8$ & $18.8$ & $8.0$ & $38.5$ & $16.7$ & $24.3$ & $58.3$ & $9.7$ & $8.7$ & $28.6$ & $38.9$ & $17.7$ & $26.1$ \\
    SFT          & $\underline{97.1}$ & $74.1$ & $\underline{87.5}$ & $52.0$ & $38.5$ & $41.7$ & $68.6$ & $\underline{91.7}$ & $45.2$ & $65.2$ & $81.0$ & $50.0$ & $35.3$ & $61.9$\\
    Vanilla OPD & $\mathbf{100.0}$ & $\mathbf{96.3}$ & $\underline{87.5}$ & $84.0$ & $69.2$ & $79.2$ & $88.6$ & $\mathbf{95.8}$ & $\underline{90.3}$ & $82.6$ & $90.5$ & $55.6$ & $\underline{76.5}$ & $83.6$ \\
    \midrule
    \rowcolor{gray!8}
    \multicolumn{15}{l}{\textit{Reference-Augmented OPD}} \\
    Random               & $\mathbf{100.0}$ & $\mathbf{96.3}$ & $\mathbf{93.8}$ & $\underline{88.0}$ & $69.2$ & $83.3$ & $\underline{90.7}$ & $\mathbf{95.8}$ & $\underline{90.3}$ & $91.3$ & $\underline{95.2}$ & $44.4$ & $70.6$ & $83.6$ \\
    Shortest             & $\mathbf{100.0}$ & $85.2$ & $\mathbf{93.8}$ & $\mathbf{92.0}$ & $53.9$ & $70.8$ & $85.7$ & $87.5$ & $80.7$ & $\mathbf{100.0}$ & $\mathbf{100.0}$ & $\underline{77.8}$ & $\underline{76.5}$ & $87.3$ \\
    BM25                 & $\mathbf{100.0}$ & $85.2$ & $\mathbf{93.8}$ & $\underline{88.0}$ & $76.9$ & $\mathbf{91.7}$ & $\underline{90.7}$ & $\underline{91.7}$ & $\underline{90.3}$ & $\mathbf{100.0}$ & $\underline{95.2}$ & $66.7$ & $\mathbf{88.2}$ & $89.6$ \\
    Qwen3-Embedding-0.6B & $\mathbf{100.0}$ & $\mathbf{96.3}$ & $75.0$ & $80.0$ & $\mathbf{100.0}$ & $83.3$ & $90.0$ & $87.5$ & $\underline{90.3}$ & $\underline{95.7}$ & $\mathbf{100.0}$ & $\mathbf{88.9}$ & $\underline{76.5}$ & $\underline{90.3}$\\
    Edit Distance        & $\underline{97.1}$ & $81.5$ & $\mathbf{93.8}$ & $\underline{88.0}$ & $76.9$ & $\underline{87.5}$ & $88.6$ & $\mathbf{95.8}$ & $\underline{90.3}$ & $\underline{95.7}$ & $\underline{95.2}$ & $72.2$ & $\underline{76.5}$ & $88.8$ \\
    LCP                  & $\mathbf{100.0}$ & $\mathbf{96.3}$ & $\mathbf{93.8}$ & $84.0$ & $69.2$ & $83.3$ & $90.0$ & $\underline{91.7}$ & $83.9$ & $\underline{95.7}$ & $\mathbf{100.0}$ & $50.0$ & $\underline{76.5}$ & $84.3$ \\
    BRTS & $\mathbf{100.0}$ & $\underline{92.6}$ & $\mathbf{93.8}$ & $84.0$ & $30.8$ & $\mathbf{91.7}$ & $87.1$ & $\underline{91.7}$ & $\mathbf{100.0}$ & $87.0$ & $\underline{95.2}$ & $22.2$ & $70.6$ & $81.3$ \\
    \rowcolor{methodshade}
    LCS & $\mathbf{100.0}$ & $88.9$ & $\mathbf{93.8}$ & $\underline{88.0}$ & $\underline{92.3}$ & $83.3$ & $\mathbf{91.4}$ & $\underline{91.7}$ & $\mathbf{100.0}$ & $\underline{95.7}$ & $\underline{95.2}$ & $\mathbf{88.9}$ & $\mathbf{88.2}$ & $\mathbf{94.0}$ \\
    \bottomrule
    \end{tabular}
    \end{adjustbox}
    \caption{\textbf{ALFWorld results.}
Success rates are reported by task category and overall on Valid Seen and
Valid Unseen.
The best and second-best results for each student are shown in \textbf{bold} and
\underline{underlined}, respectively. LCS is the only reference selector to
improve both overall scores over vanilla OPD for both students.
    }
\label{tab:alfworld_results}
\vspace{-0.3cm}
\end{table*}

\begin{table}[t]
    \centering
    \setlength{\tabcolsep}{8pt}
    \begin{adjustbox}{width=0.7\linewidth}
    \begin{tabular}{lllllll}
    \toprule
    & \multicolumn{3}{c}{$\tau^3$ Airline (\%)} & \multicolumn{3}{c}{$\tau^3$ Retail (\%)}\\
    \cmidrule(lr){2-4} \cmidrule(lr){5-7}
    Approach & Pass\textasciicircum 1 & Pass\textasciicircum 4 & Pass@4 & Pass\textasciicircum 1 & Pass\textasciicircum 4 & Pass@4 \\
    \midrule
    Teacher Zero-Shot    & $82.9$ & $51.7$ & $96.7$ & $70.6$ & $37.5$ & $95.0$ \\
    \midrule
    \rowcolor{gray!15}
    \multicolumn{7}{c}{
        \textbf{Qwen3.5-2B} \textcolor{gray}{(Student)}
    } \\
    Zero-Shot    & $21.3$ & $0.0$  & $53.3$ & $50.6$ & $14.2$ & $85.8$ \\
    SFT & $10.4$ & $0.0$ & $26.7$ & $20.0$  & $1.7$  & $52.5$  \\
    Vanilla OPD          & $64.6$ & $30.0$ & $86.7$ & $70.4$ & $38.3$ & $92.5$ \\
    \midrule
    \rowcolor{gray!8}
    \multicolumn{7}{l}{\textit{Reference-Augmented OPD}} \\
    Random               & $64.6$ & $31.7$ & $83.3$ & $73.1$ & $40.8$ & $94.2$ \\
    Shortest & $59.2$ & $21.7$ & $85.0$ & $70.2$ & $40.0$ & $\underline{95.0}$ \\
    BM25                 & $\mathbf{69.6}$ & $\underline{35.0}$ & $\mathbf{91.7}$ & $75.6$ & $48.3$ & $92.5$ \\
    Qwen3-Embedding-0.6B & $65.0$ & $33.3$ & $88.3$ & $\underline{76.5}$ & $\mathbf{52.5}$ & $94.2$ \\
    Edit Distance        & $60.8$ & $25.0$ & $88.3$ & $71.9$ & $39.2$ & $94.2$ \\
    LCP                  & $54.2$ & $16.7$ & $85.0$ & $75.0$ & $44.2$ & $94.2$ \\
    BRTS                 & $60.0$ & $23.3$ & $85.0$ & $73.3$ & $44.2$ & $94.2$ \\
    \rowcolor{methodshade}
    LCS & $\underline{68.8}$ & $\mathbf{41.7}$ & $\underline{90.0}$ & $\mathbf{78.3}$ & $\underline{50.8}$ & $\mathbf{95.8}$ \\
    \midrule 
    \rowcolor{gray!15}
    \multicolumn{7}{c}{
        \textbf{Qwen3.5-4B} \textcolor{gray}{(Student)}
    } \\
    Zero-Shot    & $13.3$ & $0.0$ & $31.7$ & $47.5$ & $10.0$ & $80.0$\\
    SFT          & $28.3$ & $1.7$ & $68.3$ & $22.5$ & $0.8$ & $60.0$ \\
    Vanilla OPD  & $\underline{51.2}$ & $\underline{26.7}$ & $\underline{71.7}$ & $56.2$ & $20.0$ & $86.7$ \\
    \midrule
    \rowcolor{gray!8}
    \multicolumn{7}{l}{\textit{Reference-Augmented OPD}} \\
    Random               & $35.0$ & $15.0$ & $60.0$ & $55.2$ & $\underline{24.2}$ & $85.8$ \\
    Shortest & $43.3$ & $21.7$ & $68.3$ & $54.8$ & $21.7$ & $87.5$ \\
    BM25 & $23.3$ & $6.7$ & $43.3$ & $52.1$ & $18.3$ & $86.7$ \\
    Qwen3-Embedding-0.6B & $30.4$ & $10.0$ & $55.0$ & $\underline{56.7}$ & $20.0$ & $\underline{89.2}$\\
    Edit Distance        & $27.1$ & $5.0$ & $50.0$ & $43.5$ & $14.2$ & $73.3$ \\
    LCP & $27.1$ & $5.0$ & $51.7$ & $45.4$ & $6.7$ & $80.8$\\
    BRTS & $28.7$ & $8.3$ & $53.3$ & $38.7$ & $11.7$ & $66.7$ \\
    \rowcolor{methodshade}
    LCS & $\mathbf{65.4}$ & $\mathbf{38.3}$ & $\mathbf{93.3}$ & $\mathbf{69.4}$ & $\mathbf{35.0}$ & $\mathbf{92.5}$\\
    \bottomrule
    \end{tabular}
    \end{adjustbox}
    \caption{\textbf{$\tau^3$-bench results.}
Airline is evaluated in-domain; Retail tests cross-domain transfer without further optimization.
Best and second-best results per student are \textbf{bold} and \underline{underlined}, respectively. LCS improves vanilla OPD across all six metrics. 
    }
    \label{tab:tau_results}
    \vspace{-0.3cm}
\end{table}

\subsection{Experiment Setup}

\textbf{Benchmarks and Models.} We evaluate on the embodied household environment
ALFWorld~\citep{shridhar2021alfworld} and the tool-use customer-service
benchmark $\tau^3$-bench~\citep{yao2024tau}. 
We follow benchmark splits, using training tasks for reference collection and optimization and held-out tasks only for evaluation. For ALFWorld, we evaluate two student models, Qwen2.5-3B-Instruct and Qwen2.5-7B-Instruct~\citep{qwen2025qwen25technicalreport}, using a GiGPO-trained Qwen2.5-7B-Instruct~\citep{feng2025group} as the teacher. For $\tau^3$-bench, we use Qwen3.5-2B and Qwen3.5-4B as student models, Qwen3.5-9B~\citep{qwen3.5} as the teacher, and Claude Opus 4.7 as the simulated user. Prompts and training/inference settings appear in Appendix~\ref{appendix:prompt} and~\ref{appendix:experiment_setting}.

\textbf{Evaluation Protocol.} For ALFWorld, we report overall and per-category success on Valid Seen (new task instances in training rooms) and Valid Unseen (unseen rooms and layouts).
For $\tau^3$-bench, we collect references and
optimize only on the airline domain, evaluate in-domain on held-out test airline
tasks, and use the retail domain solely for out-of-domain evaluation. At
evaluation time, we run four independent trials per task. Following
$\tau$-bench~\citep{yao2024tau}, Pass\textasciicircum 1 is the average single-trial
success rate, Pass\textasciicircum 4 is the fraction of tasks solved in all four
trials, and Pass@4 is the fraction solved in at least one trial. To
reduce run-to-run variance from the relatively small task sets and
stochastic interactions, all $\tau^3$-bench results are averaged over
three random seeds.

\textbf{Baselines.} We compare with zero-shot students and teachers, supervised fine-tuning (SFT), and vanilla OPD. Reference-augmented OPD methods share a pool of reward-validated successes from four teacher rollouts per training task.
We evaluate seven alternative reference-selection strategies. \textbf{Random} uniformly samples a candidate reference. \textbf{Shortest} selects the candidate with the fewest turns, 
following the shortest-trajectory criterion of~\citet{penaloza2026privileged}. \textbf{BM25}~\citep{conf/trec/RobertsonWJHG94,10.1561/1500000019}
and \textbf{Qwen3-Embedding-0.6B}~\citep{qwen3embedding} rank by lexical and semantic similarity to the failed trajectory, respectively. Among action-sequence-based alternatives, \textbf{Edit Distance} selects the candidate with the smallest
Levenshtein distance, which measures the cost of transforming one action sequence into another through insertions, deletions, and substitutions, whereas \textbf{LCP} selects the candidate sharing the longest common action prefix. Finally, following the rollout-selection criterion of
\textbf{BRTS}~\citep{zhang2026on}, we select the candidate with the highest token-level overlap with the student's top-$K$ candidate set. Vanilla OPD updates on all rollouts, whereas all reference-augmented variants use failure-only updates.

\subsection{Results and Analyses}

\textbf{Effectiveness of LCS.} Our LCS-based method yields the most consistent gains across the primary evaluations. On ALFWorld (Table~\ref{tab:alfworld_results}), it achieves the highest overall success rates on both splits for each student. On the more challenging Valid Unseen split, LCS improves overall success over vanilla OPD from $78.4\%$ to $85.8\%$ for Qwen2.5-3B-Instruct and from $83.6\%$ to $94.0\%$ for Qwen2.5-7B-Instruct. On $\tau^3$-bench airline (Table~\ref{tab:tau_results}), it improves all three metrics for both students and achieves the highest student Pass\textasciicircum 4, increasing it from $30.0\%$ to $41.7\%$ for Qwen3.5-2B and from $26.7\%$ to $38.3\%$ for Qwen3.5-4B. Although BM25 is slightly higher on Pass\textasciicircum 1 and Pass@4 for the 2B
student, LCS ranks first on all three metrics for the
4B student. For Qwen3.5-4B, all non-LCS selectors underperform vanilla OPD across all three airline metrics. The model's particularly low zero-shot success rate $(13.3 \%)$ may make reference-augmented distillation more sensitive to reference selection, as poorly aligned references can provide misleading guidance when the student's initial behavior is unreliable.
Detailed category-level analysis for ALFWorld is in Appendix~\ref{appendix:alfworld_per_category}.

\textbf{Comparison of Alignment Criteria.} Across the aggregate metrics in Tables~\ref{tab:alfworld_results} and~\ref{tab:tau_results}, LCS consistently outperforms both Edit Distance and LCP. Although Edit Distance is sensitive to action order, it reduces trajectory comparison to a global transformation cost rather than directly measuring preserved ordered action structure, while LCP discards all agreement after the first mismatch. LCS instead directly identifies the longest shared ordered action backbone, emphasizing sequential compatibility while tolerating unmatched actions that may correspond to recoverable detours. Together, these comparisons support preserved ordered action agreement, rather than global sequence transformation cost or exact prefix agreement, as an effective criterion for selecting compatible references.

\begin{figure*}[t]
    \centering
    \includegraphics[width=1.0\linewidth]{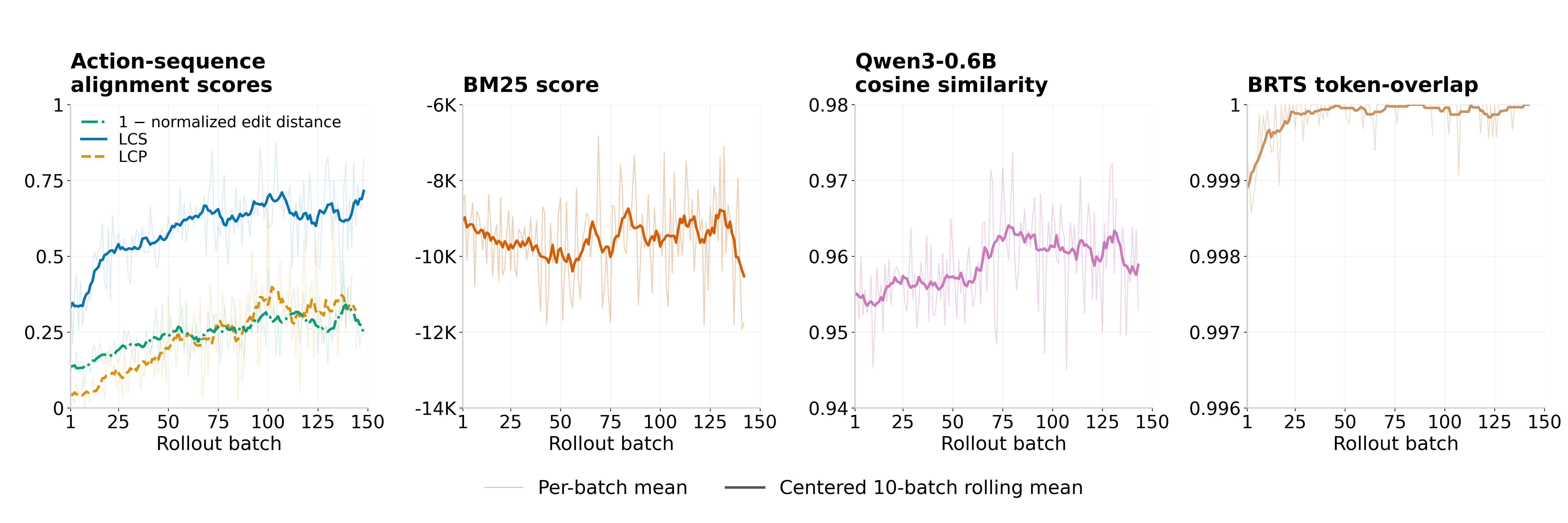}
    \vspace{-0.8cm}
    \caption{\textbf{Reference-selection score dynamics over failed ALFWorld rollouts during OPD training (student: Qwen2.5-3B-Instruct).} Rising LCS scores correspond to decreasing nearest-reference assignment costs. Normalized LCS remains substantially above normalized LCP and 1 - normalized edit distance. BM25 and Qwen3-0.6B cosine similarity exhibit noisy dynamics, with the latter additionally compressed within a narrow high-similarity range, while BRTS overlap rapidly saturates near one. See Appendix~\ref{appendix:score_dynamics} for an explanation of the BM25 score scale.
    }
    \label{fig:failure_rollout_selector_scores}
\end{figure*}

\begin{table*}[t]
    \centering
    \begin{adjustbox}{width=\linewidth}
    \begin{tabular}{llllllllll}
    \toprule
    & Success & \multicolumn{2}{c}{ALFWorld Success Rate (\%)} & \multicolumn{3}{c}{$\tau^3$ Airline (\%)} & \multicolumn{3}{c}{$\tau^3$ Retail (\%)}\\
    \cmidrule(lr){3-4} \cmidrule(lr){5-7} \cmidrule(lr){8-10}
    Approach & Rollout & Valid Seen & Valid Unseen & Pass\textasciicircum 1 & Pass\textasciicircum 4 & Pass@4 & Pass\textasciicircum 1 & Pass\textasciicircum 4 & Pass@4 \\
    \midrule
    Vanilla OPD                  & \checkmark & $79.3$ & $\mathbf{78.4}$ & $\mathbf{64.6}$ & $30.0$ & $\mathbf{86.7}$ & $70.4$ & $38.3$ & $92.5$ \\
                         &            & 
    $\mathbf{81.4}$ & $71.6$ & $62.9$ & $\mathbf{31.7}$ & $\mathbf{86.7}$ & $\mathbf{71.9}$ & $\mathbf{44.2}$ & $\mathbf{93.3}$\\
    \midrule
    LCS & \checkmark & $69.3$ & $72.4$ & $57.1$ & $21.7$ & $83.3$ & $68.5$ & $37.5$ & $93.3$\\
    & & $\mathbf{85.7}$ & $\mathbf{85.8}$ & $\mathbf{68.8}$ & $\mathbf{41.7}$ & $\mathbf{90.0}$ & $\mathbf{78.3}$ & $\mathbf{50.8}$ & $\mathbf{95.8}$ \\
    \bottomrule
    \end{tabular}
    \end{adjustbox}
    \caption{Correcting already successful rollouts consistently degrades LCS-based OPD performance, evaluated using Qwen2.5-3B-Instruct on ALFWorld and Qwen3.5-2B on \(\tau^3\)-bench. Better results are \textbf{bolded}.
    }
     \label{tab:fail_ablation}
      \vspace{-0.2cm}
\end{table*}

\textbf{Selector-Score Dynamics.} Fig.~\ref{fig:failure_rollout_selector_scores} compares reference-selector score dynamics on failed rollouts used in ALFWorld training, which form the empirical source distribution in our OT formulation. Normalized LCS is substantially higher than normalized LCP from the beginning and remains higher than it throughout training. The early gap between LCS and LCP indicates that many failures retain considerable ordered action agreement with a successful trajectory after an early divergence. LCS rises through training before reaching a noisy, non-saturated plateau, suggesting that even failed rollouts become increasingly compatible with their nearest successful references, which corresponds to a decreasing mean assignment cost.
Unlike LCS, 1 - normalized edit distance improves modestly, BM25 shows no consistent upward trend, Qwen3-Embedding-0.6B cosine similarity fluctuates within a narrow high-similarity range, and BRTS overlap rapidly saturates near one. 

Fig.~\ref{fig:all_rollout_selector_scores} in Appendix shows the complementary aggregate view over all student rollouts from the same ALFWorld OPD training run. Selected-reference scores generally increase across criteria, indicating broad movement toward successful reference trajectories. LCS rises more strongly than in the failure-only view and exhibits smaller batch-to-batch fluctuations than several alternatives, while the other criteria show patterns consistent with Fig.~\ref{fig:failure_rollout_selector_scores}. 

\textbf{Generalizability.} We evaluate cross-domain transfer by applying the airline-trained
checkpoints directly to $\tau^3$-bench retail without further optimization (Table~\ref{tab:tau_results}). LCS improves all three metrics over vanilla OPD for both students. Pass\textasciicircum 1 increases from $70.4\%$ to $78.3\%$ for Qwen3.5-2B and from
$56.2\%$ to $69.4\%$ for Qwen3.5-4B. LCS also achieves the highest student Pass\textasciicircum 1 and Pass@4 for both models. Together with its strong performance on ALFWorld Valid Unseen, these results indicate that compatibility-aware reference selection remains effective under both environment-level and cross-domain distribution shifts.

\textbf{Failure-Only Updates Benefit LCS-Based OPD.} Table~\ref{tab:fail_ablation} disentangles the effects of failure-only training and LCS-based reference selection for smaller students, with larger-student results in Table~\ref{tab:fail_ablation_group_2} (Appendix~\ref{appendix:failure-only}). Across all four students, including successful rollouts has mixed effects for vanilla OPD but consistently degrades LCS-based reference-augmented OPD across all reported metrics. These results support the correction view that reward-validated successes already realize acceptable behavior, such that further teacher supervision may introduce unnecessary or erroneous updates that disrupt valid student behavior.

\section{Related Work}

\textbf{Privileged Supervision.}
Recent work augments distillation with information unavailable to the student at decision time, including post-action environment feedback, peer rollouts, reference solutions,
and expert demonstrations~\citep{hubotter2026reinforcement,yu2026multi,li2026distill, zhao2026selfdistilled, shenfeld2026selfdistillation, yu2026failure}. However, privileged solution conditioning can bias supervision toward a particular solution rather than preserve alternative correct behavior~\citep{harne2026privileged}. In mathematical reasoning, step-aligned critiques provide more targeted supervision than complete reference solution~\citep{kara2026role}. In multi-turn agentic settings, SMRC-SD uses environment-specific state signatures to determine whether a given successful reference
supports the student's current execution state, routing and
contextualizing reference-conditioned supervision at matched
turns~\citep{liu2026privileged}.
Our work addresses a complementary decision: which successful
reference should condition the teacher for a particular failed
rollout?

\textbf{Reference Selection.}
For mathematical reasoning, BRTS selects teacher
rollouts by considering token-level alignment
with the student~\citep{zhang2026on}.
In multi-turn agentic settings, \citet{penaloza2026privileged}
select the successful trajectory with the fewest steps for each
task to construct privileged information. We align each reference with its failed rollout using normalized LCS, preserving ordered action agreement despite unmatched actions as a proxy for latent state-path compatibility under partial observability without explicit state reconstruction. Broader OPD and multi-turn OPD methods are reviewed in Appendix~\ref{appendix:related}.
\section{Conclusion}

We study which successful trajectories provide useful reference for multi-turn OPD. We formulate reference selection as free-target optimal transport and instantiate compatibility with normalized LCS over action sequences, preserving ordered agreement while tolerating recoverable detours. Across ALFWorld and $\tau^3$-bench airline, LCS is the only evaluated selector to improve over reference-free OPD on both ALFWorld overall scores and all three $\tau^3$-bench airline metrics. The airline-trained model also transfers to the retail domain of $\tau^3$-bench without further optimization. These results establish trajectory compatibility as a central criterion for reference selection in multi-turn OPD.

\bibliography{iclr2026_conference}
\bibliographystyle{iclr2026_conference}

\appendix

\clearpage

\section{Pseudocode}
\label{appendix:pseudocode}

See Algorithm~\ref{alg:lcs-ref-opd}.

\begin{algorithm}[t]
\caption{Failure-Only OPD with LCS Reference Selection}
\label{alg:lcs-ref-opd}
\begin{algorithmic}[1]
\Require Training tasks $\mathcal{D}$, teacher policy $\pi_\phi$,
student policy $\pi_\theta$, success evaluator $\operatorname{Succ}$,
fixed offline rollout count $K$, batch size $B$

\Statex \textbf{Offline reference collection (once)}
\State For each $x\in\mathcal{D}$, draw $K$ teacher rollouts and
retain the reward-validated successes as $\mathcal{R}_x$

\Statex \textbf{Online OPD training step}
\State Sample a minibatch of task instances
$\{x_i\}_{i=1}^{B}$ from $\mathcal{D}$
\State For each $x_i$, sample a student rollout
$\tau_i\sim\pi_\theta(\cdot\mid x_i)$
\State Retain the failed rollouts:
$\mathcal{I}^- \gets
\{i:\operatorname{Succ}(x_i,\tau_i)=0\}$

\ForAll{$i\in\mathcal{I}^-$}
    \If{$\mathcal{R}_{x_i}\neq\varnothing$}
        \State Select
        $\displaystyle
        \tau_i^*
        \in
        \arg\max_{\tau^+\in\mathcal{R}_{x_i}}
        s_{\mathrm{LCS}}(\tau_i,\tau^+)
        $
        \State Compute the OPD loss $\mathcal{L}_i$ on $\tau_i$
        with the teacher conditioned on $\tau_i^*$
    \Else
        \State Compute the standard OPD loss $\mathcal{L}_i$ on
        $\tau_i$ without a reference
    \EndIf
\EndFor

\If{$\mathcal{I}^-\neq\varnothing$}
    \State Update the student using
    $\displaystyle
    \frac{1}{|\mathcal{I}^-|}
    \sum_{i\in\mathcal{I}^-}\mathcal{L}_i
    $
\EndIf
\end{algorithmic}
\end{algorithm}

\section{Proofs of Theoretical Results}
\label{appendix:proof}

\subsection{Proof of Proposition~\ref{prop:optimal_reference_assignment}}
\label{appendix:proof-prop-optimal-reference}

\noindent\textbf{
Proposition~\ref{prop:optimal_reference_assignment}
(Optimal Reference Assignment; Full Statement).}
\textit{Suppose that $\mathcal{R}_x$ is finite and nonempty, and enumerate its
elements as
$
    \mathcal{R}_x
    =
    \{\tau_1^+,\ldots,\tau_{M_x}^+\}
$. Suppose also that
$c(\tau^-,\tau^+)\in\mathbb{R}$ for every failure--success pair. Then the
free-target optimal transport objective in
Equation~\ref{eq:free_target_ot} satisfies
\begin{equation*}
    \mathsf{OT}_{c}(\mu_x^-,\mathcal{R}_x)
    =
    \frac{1}{N_x}
    \sum_{i=1}^{N_x}
    \min_{\tau^+\in\mathcal{R}_x}
    c(\tau_i^-,\tau^+).
\end{equation*}
Moreover, writing $\gamma_{ij}$ for the transport mass assigned from
$\tau_i^-$ to $\tau_j^+$, there exists an optimal deterministic transport
plan satisfying
\begin{equation*}
    \gamma_{ij}^*
    =
    \frac{1}{N_x}
    \mathbb{I}\{j=j_i^*\},
    j_i^*
    \in
    \arg\min_{1\leq j\leq M_x}
    c(\tau_i^-,\tau_j^+).
\end{equation*}
}

\begin{proof}
The source-marginal constraint requires
\begin{equation}
    \sum_{j=1}^{M_x}\gamma_{ij}
    =
    \frac{1}{N_x}
    \qquad
    \text{for each } i.
\end{equation}

Define $q_{ij}=N_x\gamma_{ij}$. Then $q_{ij}\geq 0$ and
$\sum_{j=1}^{M_x}q_{ij}=1$, so $(q_{i1},\ldots,q_{i M_x})$ specifies a
distribution over the candidate references for the $i$-th failed rollout.
Because the target marginal is unconstrained, the column sums of
$\gamma$ simply determine $\nu$, so there are no additional constraints
coupling the choices for different failed rollouts. The transport objective therefore becomes
\begin{equation}
    \frac{1}{N_x}
    \sum_{i=1}^{N_x}
    \min_{\substack{q_{ij}\geq 0\\
                    \sum_{j=1}^{M_x}q_{ij}=1}}
    \sum_{j=1}^{M_x}
    q_{ij}\,
    c(\tau_i^-,\tau_j^+).
\end{equation}
Each inner objective is minimized by placing all mass on a minimum-cost
reference. Hence,
\begin{equation}
    \mathsf{OT}_c(\mu_x^-,\mathcal{R}_x)
    =
    \frac{1}{N_x}
    \sum_{i=1}^{N_x}
    \min_{\tau^+\in\mathcal{R}_x}
    c(\tau_i^-,\tau^+),
\end{equation}
and there exists an optimal deterministic plan that assigns each failed
rollout to one of its minimum-cost references.
\end{proof}

\subsection{Proof of Theorem~\ref{thm:lcs_recoverable_detours}}
\label{appendix:proof_lcs_recoverable_detours}

To characterize when state agreement is nevertheless recovered, consider
the unmatched action blocks preceding each matched pair. Let
$i_0=j_0=-1$ and define potentially empty unmatched action blocks:
\begin{align*}
B_r^-
&=
(a_{i_{r-1}+1}^-,\ldots,a_{i_r-1}^-),
\\ 
B_r^+
&=
(a_{j_{r-1}+1}^+,\ldots,a_{j_r-1}^+),
\qquad r=1,\ldots,m,
\label{eq:lcs_unmatched_blocks}
\end{align*}
For an action block $B$, let $F_B(s)$ denote the state obtained by sequentially applying the transition function $F$ starting from $s$,
with $F_{\emptyset}(s)=s$. We call the unmatched blocks preceding a matched pair \emph{recoverable}
if, whenever they are entered from the same state, they reconverge to the
same state before the matched action.

\noindent\textbf{
Theorem~\ref{thm:lcs_recoverable_detours}
(State Consistency under Recoverable Detours).}
\textit{
Suppose $\tau^-$ and $\tau^+$ share the same initial state and follow the same deterministic transition function $F$. Suppose further
that, before every matched pair $(i_r,j_r)$, the unmatched blocks $B_r^-$ and $B_r^+$ are recoverable: for any common entry state $s$ encountered by the two trajectories,
\begin{equation}
F_{B_r^-}(s)=F_{B_r^+}(s).
\label{eq:recoverable_detour}
\end{equation}
Then every matched pair in
$A_{\mathrm{LCS}}(\tau^-,\tau^+)$ is state-consistent:
\begin{equation*}
s_{i_r}^-=s_{j_r}^+,
\qquad r=1,\ldots,m.
\label{eq:lcs_state_consistency}
\end{equation*}
}

\begin{proof}
If $m=0$, the claim holds trivially. Otherwise, we proceed by
induction over the matched pairs.

For the first pair, the shared initial state and recoverability
of the preceding unmatched blocks give
\[
s^-_{i_1}
= F_{B^-_1}(s^-_0)
= F_{B^+_1}(s^+_0)
= s^+_{j_1}.
\]

Suppose $s^-_{i_r}=s^+_{j_r}$ for some $1\le r<m$.
Since $a^-_{i_r}=a^+_{j_r}$ and the transition is deterministic,
the states immediately after these matched actions also agree:
$s^-_{i_r+1}=s^+_{j_r+1}$.
Recoverability of the next unmatched blocks then yields
\[
s^-_{i_{r+1}}
= F_{B^-_{r+1}}(s^-_{i_r+1})
= F_{B^+_{r+1}}(s^+_{j_r+1})
= s^+_{j_{r+1}}.
\]
Thus, every matched pair is state-consistent.
\end{proof}

\subsection{Proof of Theorem~\ref{thm:reference_supervision_gain}}
\label{appendix:proof_reference_benefit}

We first derive a general bound allowing only a subset of LCS matches to be state-consistent, then recover Theorem~\ref{thm:reference_supervision_gain} as a special case.

$\Delta_t(\tau^+)>0$ means that reference conditioning assigns higher
expected log probability to desirable responses than reference-free
supervision at the same history. Let
\begin{equation*}
\mathcal I(\tau^-,\tau^+)
=
\left\{
i_r
\;\middle|\;
\substack{
(i_r,j_r)\in A_{\mathrm{LCS}}(\tau^-,\tau^+),\\
s_{i_r}^-=s_{j_r}^+
}
\right\}
\label{eq:state_consistent_lcs_turns}
\end{equation*}
denote the turns of the failed trajectory participating in state-consistent
LCS matches.

\textbf{Teacher Responsiveness.}
Suppose that, for the reference $\tau^+$ under consideration, there exist
constants $g>0$ and $b\geq0$ such that reference conditioning yields average
gain at least $g$ over the state-consistent LCS-matched turns,
\begin{equation*}
\sum_{t\in\mathcal I(\tau^-,\tau^+)}
\Delta_t(\tau^+)
\geq
g\,|\mathcal I(\tau^-,\tau^+)|,
\label{eq:teacher_gain_matched}
\end{equation*}
while its average degradation over the remaining turns is bounded by $b$,
\begin{equation*}
\sum_{t\notin\mathcal I(\tau^-,\tau^+)}
\Delta_t(\tau^+)
\geq
-b
\left(
T^- - |\mathcal I(\tau^-,\tau^+)|
\right).
\label{eq:teacher_harm_unmatched}
\end{equation*}
This condition separates structural compatibility from teacher behavior:
a compatible reference is useful only if the teacher makes productive use of
the corresponding hindsight.

\noindent\textbf{
Theorem~\ref{thm:reference_supervision_gain}
(Benefit over Reference-Free Supervision).}
\textit{
Under the conditions of
Theorem~\ref{thm:lcs_recoverable_detours}
and teacher-responsiveness, let
$m=\ell_{\mathrm{LCS}}(\tau^-,\tau^+)$.
Then
\begin{equation*}
\frac{1}{T^-}
\sum_{t=0}^{T^--1}
\Delta_t(\tau^+)
\geq
(g+b)\frac{m}{T^-}-b.
\end{equation*}
Consequently, conditioning on $\tau^+$ yields
strictly positive average supervision gain whenever
\begin{equation}
\frac{m}{T^-}>\frac{b}{g+b}.
\label{eq:positive_reference_gain}
\end{equation}
}

\begin{proof}
Partition the turns of $\tau^-$ into the state-consistent
LCS-matched turns $\mathcal{I}(\tau^-,\tau^+)$ and the remaining
turns.
Let $I=\mathcal{I}(\tau^-,\tau^+)$.
The teacher-responsiveness condition gives
\[
\begin{aligned}
\sum_{t=0}^{T^- -1}\Delta_t(\tau^+)
&= \sum_{t\in I}\Delta_t(\tau^+) \\
&\quad+
\sum_{\substack{0\le t<T^-\\t\notin I}}
\Delta_t(\tau^+) \\
&\ge g|I|-b\bigl(T^- -|I|\bigr).
\end{aligned}
\]
Dividing by $T^- > 0$ and rearranging yields
\[
\frac{1}{T^-}
\sum_{t=0}^{T^- - 1} \Delta_t(\tau^+)
\ge
(g+b)\frac{|I|}{T^-}-b.
\]

Under the conditions of
Theorem~\ref{thm:lcs_recoverable_detours},
every LCS match is state-consistent, so $|I|=m$.
Substituting yields
\[
\frac{1}{T^-}
\sum_{t=0}^{T^- - 1}\Delta_t(\tau^+)
\ge
(g+b)\frac{m}{T^-}-b.
\]
Since $g+b>0$, this lower bound is strictly
positive whenever $m/T^->b/(g+b)$.
By the definition of $\Delta_t(\tau^+)$,
a positive average gain means that the
reference-conditioned teacher assigns higher
expected log probability to desirable responses,
averaged over turns, than the reference-free teacher.
\end{proof}

\section{Prompt Templates}
\label{appendix:prompt}

Throughout this appendix, text following // denotes explanatory comments added for clarity and is not part of the actual input passed to the models. Text enclosed in braces denotes content populated at runtime.

\subsection{ALFWorld}

For ALFWorld, \{full\_interaction\_history\} denotes all preceding user and assistant messages, excluding the current user prompt shown below, and is empty on the first turn. \{action\_history\} contains only the most recent \{history\_length\} turns, following the setup of TCOD~\citep{wang2026tcod}.

\begin{tcolorbox}
[colback=gray!5!white, colframe=gray!75!black, title=Student rollout prompt (first two turns)]
\{full\_interaction\_history\} \\
\\
You are an expert agent operating in the ALFRED Embodied Environment. Your task is to: \{task\_description\} \\
\\
Your current observation is: \{current\_observation\} \\
Your admissible actions of the current situation are: [\{admissible\_actions\}]. \\
\\
Now it's your turn to take an action. \\
You should first reason step-by-step about the current situation. This reasoning process MUST be enclosed within \texttt{\textless think\textgreater{} \textless/think\textgreater} tags.
Once you've finished your reasoning, you should choose an admissible action for current step and present it within \texttt{\textless action\textgreater{} \textless/action\textgreater} tags.
\end{tcolorbox}

\begin{tcolorbox}
[colback=gray!5!white, colframe=gray!75!black, title=Student rollout prompt (subsequent turns)]
\{full\_interaction\_history\} \\
\\
You are an expert agent operating in the ALFRED Embodied Environment. Your task is to: \{task\_description\}\\
\\
Prior to this step, you have already taken \{step\_count\} step(s). Below are the most recent \{history\_length\} observations and the corresponding actions you took: \{action\_history\} \\
\\
You are now at step \{current\_step\} and your current observation is: 
\{current\_observation\}. Your admissible actions of the current situation are: [\{admissible\_actions\}]. \\
\\
Now it's your turn to take an action. \\
You should first reason step-by-step about the current situation. This reasoning process MUST be enclosed within \texttt{\textless think\textgreater{} \textless/think\textgreater} tags.
Once you've finished your reasoning, you should choose an admissible action for current step and present it within \texttt{\textless action\textgreater{} \textless/action\textgreater} tags.
\end{tcolorbox}

\begin{tcolorbox}
[colback=gray!5!white, colframe=gray!75!black, title=Reference-augmented teacher scoring prompt]
You will be shown two trajectories from the same task instance, in this order: \\
\\
1. A completed reference trajectory that successfully solved the task. It demonstrates one successful approach, but it may not be the only way to solve the task. \\
2. An ongoing trajectory for the same task instance. \\
\\
Use the completed successful trajectory as reference that may be helpful when continuing the ongoing trajectory. Because the two trajectories may have taken different actions, their states may differ. Treat the observations, action history, inventory, and admissible actions in the ongoing trajectory as authoritative for future action decisions. State information from the reference trajectory may or may not remain applicable, depending on how the two trajectories have developed. \\
\\
\texttt{\textless successful\_task\_trial\textgreater{}} \\
\{full\_successful\_reference\} \\
\texttt{\textless /successful\_task\_trial\textgreater{}} \\
\\
The completed reference trajectory ends here. The ongoing trajectory for the same task instance begins below. \\
\\
\{ongoing\_student\_context\} // Normal student prompt starts here.
\end{tcolorbox}

\subsection{$\tau^3$-bench}

\begin{tcolorbox}
[colback=gray!5!white, colframe=gray!75!black, title=Student rollout prompt]
\texttt{\textless instructions\textgreater{}} \\
You are a customer service agent that helps the user according to the \texttt{\textless policy\textgreater{}} provided below. \\
In each turn you can either: \\
- Send a message to the user. \\
- Make a tool call. \\
You cannot do both at the same time. \\
\\
Try to be helpful and always follow the policy. Always make sure you generate valid JSON only. \\
\texttt{\textless/instructions\textgreater} \\
\texttt{\textless policy\textgreater{}} \\
\{domain\_policy\} \\
\texttt{\textless /policy\textgreater{}} \\
\\
\{history\} // denotes the sequence of user, assistant, and tool messages
\end{tcolorbox}

\begin{tcolorbox}
[colback=gray!5!white, colframe=gray!75!black, title=Reference-augmented teacher scoring prompt]
\{original\_system\_prompt\} // instructions + domain policy \\
\\
In case it helps, here is a record of a past successful task trial for your reference. Use it only as contextual guidance when judging the next assistant response; the current conversation remains authoritative. \\
\\
\texttt{\textless successful\_task\_trial\textgreater{}} \\
\{transcript\} \\
\texttt{\textless /successful\_task\_trial\textgreater{}} \\
\\
End of reference. The current conversation begins below. \\
\\
\{history\} // Normal student current trajectory up to the current turn
\end{tcolorbox}

\section{Experimental Settings}
\label{appendix:experiment_setting}

\textbf{ALFWorld.} We adopt the ALFWorld sampling settings and core optimization hyperparameters of TCOD~\citep{wang2026tcod}, with the modifications detailed below. As the original teacher checkpoint was not released, we use GiGPO-trained Qwen2.5-7B-Instruct as our teacher. We increase the teacher context limit to accommodate reference-augmented scoring.

Table~\ref{tab:alfworld_context} summarizes the context and generation limits
used for ALFWorld training, evaluation, reference collection, and teacher
scoring.

\begin{table}[h]
\centering
\begin{adjustbox}{width=\linewidth}
\begin{tabular}{llll}
\toprule
\textbf{Stage}
& \textbf{Context limit}
& \textbf{Prompt limit}
& \textbf{Generated/scored response} \\
\midrule
Student training rollout
& 10,752
& 10,240
& 512 generated \\

Evaluation
& 32,768
& 28,672
& 4,096 generated \\

Reference-augmented teacher
& 32,768
& 32,256
& Existing response, $\leq 512$ \\
\bottomrule
\end{tabular}
\end{adjustbox}
\caption{ALFWorld context and generation limits. All lengths are in tokens.}
\label{tab:alfworld_context}
\end{table}

The main OPD optimization, rollout, and evaluation settings are summarized in
Table~\ref{tab:alfworld_training}.

\begin{table}[h]
\centering
\small
\begin{tabular}{@{}p{0.48\linewidth}p{0.42\linewidth}@{}}
\toprule
\textbf{Setting} & \textbf{Value} \\
\midrule
\multicolumn{2}{@{}l}{\textit{OPD training}} \\
Learning rate & $10^{-6}$ \\
Learning-rate schedule & Constant \\
Gradient-norm clipping & 1.0 \\
Configured optimizer updates & 200 \\
Global training batch & 64 turns \\
Rollout collection batch & 16 trajectories \\
Learner microbatching & Dynamic; 16,384-token budget per learner GPU \\
Student rollout sampling & Temperature 1.0 \\
Maximum episode length & 30 environment actions \\
\midrule
\multicolumn{2}{@{}l}{\textit{Evaluation / teacher reference collection}} \\
Sampling & Temperature 0.4, top-$p=1$, no top-$k$ filtering \\
Other sampling parameters
& Minimum sampling probability 0; repetition penalty 1.0;
presence and frequency penalties 0 \\
Maximum episode length & 30 environment actions \\
\bottomrule
\end{tabular}
\caption{ALFWorld training and inference settings.}
\label{tab:alfworld_training}
\end{table}

\textbf{$\bm{\tau}^3$-bench.} We enable thinking mode for both the student and teacher. We set the context size to $20,000$ tokens and use a dynamic generation budget capped at $10,000$ tokens. The main OPD optimization, rollout, and evaluation settings are
summarized in Table~\ref{tab:tau3_training_inference}.

\begin{table}[h]
    \centering
    \small
    \setlength{\tabcolsep}{4pt}
    \renewcommand{\arraystretch}{1.15}
    \begin{tabular}{@{}p{0.43\linewidth}p{0.53\linewidth}@{}}
        \toprule
        \textbf{Setting} & \textbf{Value} \\
        \midrule
        \multicolumn{2}{@{}l}{\textit{OPD training}} \\
        Learning rate & $10^{-6}$ \\
        Learning-rate schedule & Constant \\
        Gradient-norm clipping & 1.0 \\
        Configured optimizer updates & 10 \\
        Global training batch & 15 turns \\
        Learner microbatching
        & 1 turn per GPU \\
        \midrule
        \multicolumn{2}{@{}l}{\textit{Student rollout / evaluation / teacher reference collection}} \\
        Sampling & Temperature 1.0, top-$p=0.95$, top-$k=20$ \\
        Other sampling parameters
        & Minimum sampling probability 0; repetition penalty 1.0;
          presence penalty 1.5 \\
        Maximum episode length & 200 benchmark steps \\
        Max errors & 10 \\
        Max retries & 0 \\
        \midrule
        \multicolumn{2}{@{}l}{\textit{User simulation}} \\
        Maximum generated tokens & 4,096 \\
        Thinking mode & Adaptive \\
        Reasoning effort & High \\
        \bottomrule
    \end{tabular}
    \caption{$\tau^3$-bench training and inference settings.}
    \label{tab:tau3_training_inference}
\end{table}

\textbf{Reference Pool Statistics.}
Table~\ref{tab:reference_pool_distribution} summarizes
the distribution of distinct successful references
collected per training task.

\begin{table}[t]
    \centering
    \begin{adjustbox}{width=0.7\linewidth}
    \begin{tabular}{lll}
    \toprule
    \# Distinct References & \# $\tau^3$ Airline Tasks (\%) & \# ALFWorld Tasks (\%) \\
    \midrule
    0 & 1 (3.33\%)   & 150 (4.22\%) \\
    1 & 5 (16.67\%)  & 2,280 (64.17\%) \\
    2 & 6 (20.00\%)  & 692 (19.48\%) \\
    3 & 12 (40.00\%) & 324 (9.12\%) \\
    4 & 6 (20.00\%)  & 107 (3.01\%) \\
    \midrule
    \textbf{Total}
    & \textbf{30 (100\%)}
    & \textbf{3,553 (100\%)} \\
    \bottomrule
    \end{tabular}
    \end{adjustbox}
    \caption{\textbf{Distribution of successful
    references per training task.}
    Successful references are deduplicated within each
    task by their complete parsed action sequences,
    preserving action order and repeated actions.
    Percentages are computed over all training tasks
    in each benchmark.}
    \label{tab:reference_pool_distribution}
\end{table}

\textbf{Evaluation Task Statistics.}
Table~\ref{tab:evaluation_task_counts} summarizes
the number of distinct tasks in each evaluation split.

\begin{table}[t]
    \centering
    \begin{tabular}{lll}
    \toprule
    Benchmark & Evaluation Split & \# Tasks \\
    \midrule
    ALFWorld & Valid Seen   & 140 \\
             & Valid Unseen & 134 \\
    \midrule
    $\tau^3$-bench & Airline (Test) & 20 \\
                   & Retail (Test)  & 40 \\
    \bottomrule
    \end{tabular}
    \caption{Evaluation task counts.}
    \label{tab:evaluation_task_counts}
\end{table}

\textbf{LCS Normalization.}
We compute LCS over valid environment actions for ALFWorld
and tool calls for $\tau^3$-bench.
Let $T^-$ and $T^+$ denote the lengths of the corresponding
action sequences from the failed rollout and successful reference,
respectively.
We use $Z(T^-,T^+)=\min\{T^-,T^+\}$ for ALFWorld
and $Z(T^-,T^+)=T^-$ for $\tau^3$-bench.

\section{Task-Dependent Gains on ALFWorld}
\label{appendix:alfworld_per_category}

LCS outperforms vanilla OPD on Cool across both splits and on Pick2 in Valid Unseen for each student. For Pick2, Valid Unseen success increases from $58.8\%$ to $94.1\%$ for the 3B student and from $76.5\%$ to $88.2\%$ for the 7B student. Cool and Pick2 expose two forms of sequential dependency particularly relevant to reference compatibility. In Cool, the validity of later actions depends on an earlier object-state transformation, while Pick2 requires preserving progress across repeated object-level subgoals. The Pick2 example in Fig.~\ref{fig:newspaper} illustrates how intermediate inventory states constrain progress across object-level subgoals.~\citet{feng2025group} reports pronounced drops on Cool and Pick2 without step-level advantages for reinforcement learning, while~\citet{li2026distill} highlights Pick2's dependence on preserving state across object interactions. Taken together, these observations suggest that trajectory compatibility may be particularly important when effective supervision must respect intermediate state changes and accumulated subgoal progress. However, the category-level effects remain student-dependent:
Examine improves on both splits for the 7B student but declines on Valid Unseen for the 3B student.

\section{Reference-Selection Score Dynamics over All Rollouts during ALFWorld OPD Training}
\label{appendix:score_dynamics}

\begin{figure*}[h]
    \centering
    \includegraphics[width=1.0\linewidth]{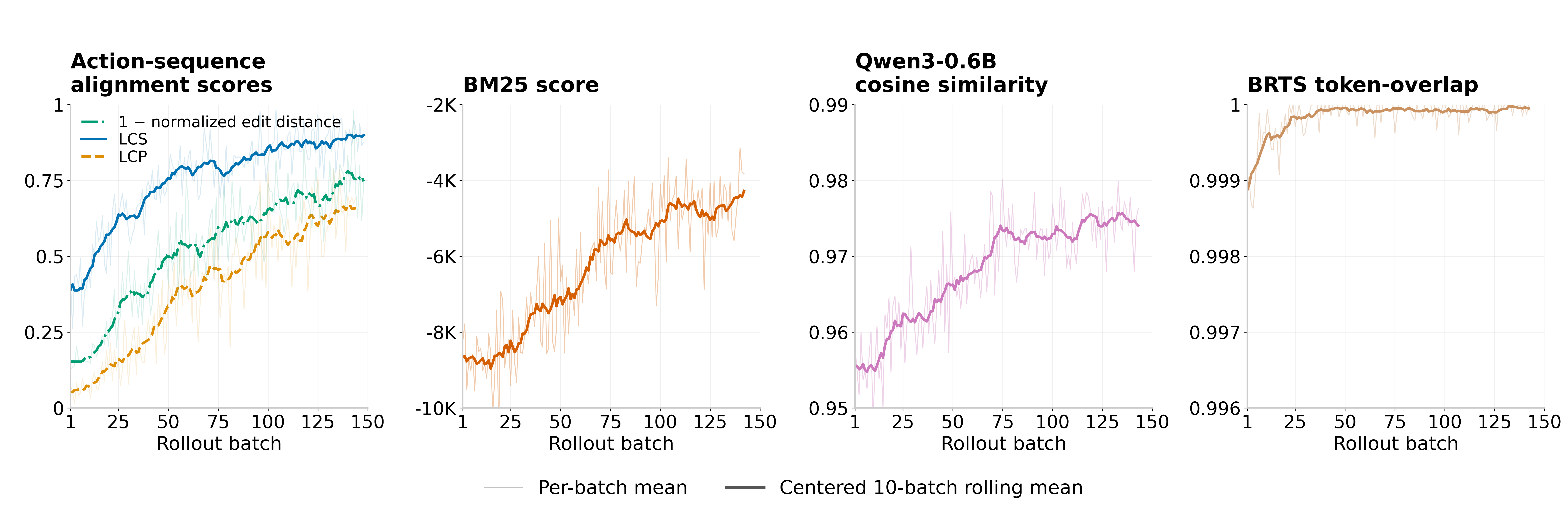}
    \vspace{-0.8cm}
    \caption{\textbf{Reference-selection score dynamics over all rollouts during ALFWorld OPD training with Qwen2.5-3B-Instruct as the student.} Selected-reference scores generally increase across criteria during training. Normalized LCS exhibits a comparatively smooth upward trend with smaller batch-to-batch fluctuations than several alternatives and remains substantially higher than normalized LCP and 1 - normalized edit distance. BM25 exhibits noisier dynamics. Qwen3-0.6B cosine similarity is compressed within a narrow high-similarity range. BRTS overlap rapidly saturates near one.}
    \label{fig:all_rollout_selector_scores}
\end{figure*}

See Fig.~\ref{fig:all_rollout_selector_scores}.

We use the BM25Okapi implementation from rank\_bm25~\citep{rank_bm25}. Each task's successful references form the retrieval corpus, and the student's failed trajectory serves as the query. In these small, task-specific pools, terms shared by most references can yield a negative mean IDF. Scores sum contributions over all query-token occurrences, including repetitions, without query-length normalization; long dialogues can therefore accumulate large negative scores.

\section{Additional Ablation Studies for Failure-Only Updates}
\label{appendix:failure-only}

Table~\ref{tab:fail_ablation_group_2} extends the failure-only ablation in Table~\ref{tab:fail_ablation} to the larger student models: Qwen2.5-7B-Instruct on ALFWorld and Qwen3.5-4B on $\tau^3$-bench. Consistent with the results for the smaller students, excluding successful rollouts from OPD updates improves LCS-based reference-augmented OPD across all reported metrics, whereas the effects on vanilla OPD are mixed.

\begin{table*}[t]
    \centering
    \begin{adjustbox}{width=\linewidth}
    \begin{tabular}{llllllllll}
    \toprule
    & Success & \multicolumn{2}{c}{ALFWorld Success Rate (\%)} & \multicolumn{3}{c}{$\tau^3$ Airline (\%)} & \multicolumn{3}{c}{$\tau^3$ Retail (\%)}\\
    \cmidrule(lr){3-4} \cmidrule(lr){5-7} \cmidrule(lr){8-10}
    Approach & Rollout & Valid Seen & Valid Unseen & Pass\textasciicircum 1 & Pass\textasciicircum 4 & Pass@4 & Pass\textasciicircum 1 & Pass\textasciicircum 4 & Pass@4 \\
    \midrule
    Vanilla OPD                  & \checkmark & $\mathbf{88.57}$ & $\mathbf{83.58}$ & $51.2$ & $\mathbf{26.7}$ & $71.7$ & $\mathbf{56.2}$ & $20.0$ & $\mathbf{86.7}$ \\
                                 &            & $85.71$ & $78.36$ & $\mathbf{54.6}$ & $23.3$ & $\mathbf{81.7}$ & $55.2$ & $\mathbf{23.3}$ & $81.7$
    \\
    \midrule
    LCS & \checkmark & $77.86$ & $76.87$ & $40.0$ & $10.0$ & $63.3$ & $60.4$ & $31.7$ & $87.5$\\
    & & $\mathbf{91.43}$ & $\mathbf{94.03}$ & $\mathbf{65.4}$ & $\mathbf{38.3}$ & $\mathbf{93.3}$ & $\mathbf{69.4}$ & $\mathbf{35.0}$ & $\mathbf{92.5}$ \\
    \bottomrule
    \end{tabular}
    \end{adjustbox}
    \caption{Correcting already successful rollouts consistently degrades LCS-based OPD performance, evaluated using Qwen2.5-7B-Instruct on ALFWorld and Qwen3.5-4B on \(\tau^3\)-bench. Better results are \textbf{bolded}.
    }
     \label{tab:fail_ablation_group_2}
\end{table*}

\section{Additional Related Work}
\label{appendix:related}

\textbf{OPD.} OPD mitigates train--inference distribution mismatch by supervising a student on its own generated sequences. GKD~\citep{agarwal2024onpolicy} unifies flexible divergence objectives and mixtures of fixed and on-policy data. MiniLLM~\citep{gu2024minillm} optimizes reverse KL to encourage the student to focus on major modes of the teacher distribution while avoiding overestimation of low-probability regions. DistiLLM~\citep{pmlr-v235-ko24c} combines skew-KL losses with adaptive off-policy reuse of student outputs, while DistiLLM-2~\citep{pmlr-v267-ko25a} introduces contrastive objectives tailored to teacher- and student-generated responses.

\textbf{Multi-Turn OPD.} In interactive environments, earlier actions alter subsequent observations, making distillation sensitive to compounding errors and the reliability of teacher supervision. TCOD~\citep{wang2026tcod} progressively expands the trajectory depth exposed during training, while TurnOPD~\citep{zhou2026turn} adaptively budgets rollout depth and progressively shifts KL loss allocation toward turn-balanced supervision. Guided-OPD~\citep{li2026onpolicy} mixes teacher- and student-generated turns under a decaying intervention schedule. SOD~\citep{zhong2026sod} reweights step-wise distillation based on student--teacher divergence, and SAGE-OPD~\citep{zhou2026sageopd} uses environment feedback and teacher judgments to selectively intervene on turns, while weighting token-level distillation by teacher confidence. ReOPD~\citep{liao2026multi} reuses teacher-collected trajectories as replayed prefixes, avoiding fresh environment interactions while balancing student-distribution relevance against teacher-target reliability. These approaches regulate rollout collection or supervision allocation, complementing the successful-reference selection studied here.

\section{Implementation Details}

We adapt the official implementation of TCOD~\citep{wang2026tcod}. The implementation is based on Trinity-RFT~\citep{pan2025trinity}, veRL~\citep{sheng2025hybridflow}, Ray~\citep{moritz2017ray}, PyTorch~\citep{NEURIPS2019_bdbca288}, and vLLM~\citep{kwon2023efficient}.

\section{Compute Resource Disclosure}

We use 8 80G A100 GPUs for OPD training. The GPUs are available via a cloud provider. 4 GPUs are used for training, 2 for teacher scoring, and 2 for model update. Training for ALFWorld takes about 10 hours while training for tau 3 airline takes about 1 hour. Across the reported training and evaluation runs, we estimate a computational budget of approximately 4,500 A100 GPU-hours. The instance we use has 1 TB of storage.

\end{document}